# Neuronal Spectral Analysis of EEG and Expert Knowledge Integration for Automatic Classification of Sleep Stages

Nizar KERKENI[(1,2)], Frédéric ALEXANDRE[(1)], Mohamed Hédi BEDOUI[(2)], Laurent BOUGRAIN[(1)], Mohamed DOGUI[(3)]
(1) Cortex Team, LORIA, BP 239, F-54506 Vandoeuvre, FRANCE
(2) TIM Team, Biophysics Laboratory, Medicine Faculty of Monastir, 5019 Monastir, TUNISIA
(3) Service of Functional Exploration of the Nervous System, CHU Sahloul, 4054 Sousse, TUNISIA
nizar.kerkeni@loria.fr   http://cortex.loria.fr/

*Abstract:* - Being able to analyze and interpret signal coming from electroencephalogram (EEG) recording can be of high interest for many applications including medical diagnosis and Brain-Computer Interfaces. Indeed, human experts are today able to extract from this signal many hints related to physiological as well as cognitive states of the recorded subject and it would be very interesting to perform such task automatically but today no completely automatic system exists. In previous studies, we have compared human expertise and automatic processing tools, including artificial neural networks (ANN), to better understand the competences of each and determine which are the difficult aspects to integrate in a fully automatic system. In this paper, we bring more elements to that study in reporting the main results of a practical experiment which was carried out in an hospital for sleep pathology study. An EEG recording was studied and labeled by a human expert and an ANN. We describe here the characteristics of the experiment, both human and neuronal procedure of analysis, compare their performances and point out the main limitations which arise from this study.

## 1 Introduction

Physiological signals, particularly including electroencephalogram (EEG) signals, can be easily recorded from the human body and their analysis can lead to identify and recognize such elements as vigilance states, pathologies of sleep or other more cognitive phenomena related to Brain-Computer Interface (BCI). Designing automatic tools to interpret such signals is of high interest but has no satisfactory solution today. In [1], we have tried to better understand the reasons why this task is so difficult, comparing human expertise and automatic analysis. On one hand, we have shown that they are difficult to compare because they do not work exactly on the same data. On the other hand, we have underlined the multi-level and multi-criteria characteristics of human expertise, difficult to reproduce with classical data analysis tools. Concerning this latter point, we have proposed that Artificial Neural Networks (ANN) technique could be a good candidate for such a task, since ANN can be used for signal processing, multi-modal analysis as well as decision-making. We are currently working at these various levels with ANN. Concerning the former point, it was important for us to compare human and ANN performances in similar conditions, which is rarely the case as stated in [1]. This is the topic of this paper in the framework of a sleep analysis study. We have chosen and will describe here sleep analysis because it is carried out in controlled medical environment and guarantees the quality of the data and of the human knowledge. Nevertheless, it is clear that this work is intended to be applicable to other more technological frameworks, like BCI.

## 2 Sleep Analysis

In the clinical routine, the study of the sleep consists of the acquisition and the recording, during one sleep night, of a physiological signals set (polysomnography), followed by a visual analysis performed by a well-trained medical expert to establish the diagnosis. This study is mainly based on three signals: electroencephalogram (EEG), electrooculogram (EOG) and electromyogram (EMG). Even if many information is lost, EEG signal reflects the cerebral electric activity. It is recorded with electrodes set on the scalp. The site of the electrodes is defined according to the nomenclature of the system 10-20 adopted by the majority of the clinical neurophysiology laboratories and is illustrated in Fig.1. EOG signal represents the ocular activity and is recorded with electrodes placed on the circumference of the eyes orbits. EMG signal corresponds to the muscular activity recorded by an electrode placed on the chin [2].

## 2.1 Visual Analysis

The visual analysis performed by the human expert consists in detecting the variations of the EEG, the EOG and the EMG during the night. These changes define the vigilance states which are the awakening and the five stages of the sleep: stage 1, stage 2, stage 3, stage 4 and the rapid eye movement sleep (REM) [3]. Basically, the visual analysis relies on the fact that each stage is characterized by the presence of one or more indicators corresponding to elementary activities and several graphical elements in the recorded signals. According to these indicators and by observing the standard rules defined by Rechtschaffen and Kales [4] the clinician learns to associate to one epoch (defined as a temporal unit used as reference, generally of 30s) a label corresponding to the estimated physiological state. The grouping-together of the all night sleep labels constitutes the hypnogram. It is a graphical representation of the various sleep stages organization during the night and makes it possible to have an compact and overall representation of the sleep architecture. The obtained hypnogram and the statistical data computed on the sleep stages will constitute a report which will be the base of the clinical decision-making.

The difficulty of the visual analysis lays in several aspects. Firstly, the quality of the signals recorded depends on the quality of the electrodes used and their setting (on the scalp or elsewhere). A bad setting can produce noise and artifacts, due to bad contacts or patient movements and complicate the signal interpretation. Secondly, the correctness of the rules used for visual interpretation themselves can be evoked. Indeed, these classical rules are based on the visual detection of some particular waves and some graphical elements present in the physiological signals and in particular in the EEG, as described in Table 1. In addition to the Alpha, Theta and Delta waves, main EEG indicators to discriminate vigilance states, the bands of sleep also gather the Sigma waves (from 12 to 16 Hz), including sleep spindles, and the Beta waves (from 16 to 32 Hz). Identifying these characteristics requires a long training by the human expert and it is observed that subjective interpretation of these rules remains possible.

| Indicator | Definition | State |
|---|---|---|
| Alpha wave | Frequency of 8 to 12Hz | Awake |
| Theta wave | Frequency of 4 to 8Hz | Stage 1, 2 and REM |
| Delta wave | Frequency of 0.5 to 4 Hz | Stage 3 and 4 |
| K complex | Transient slow waves | Stage 2 |
| Spindles | Frequency of 12 to 14 Hz | Stage 2 |
| Vertex sharp wave | Pointed waves with great amplitude | Stage 1 |
| Sawtooth | Saw tooth pattern | REM |

**Table 1**: main EEG indicators.

In fact, the detection of these indicators is more or less difficult depending on the clinician and its experience. Moreover it is frequent that these elements overlap or are affected by artifacts. The rules of visual analysis include a part of subjectivity, for example the presence of the Delta waves in more than 50% of the time period, which can lead to discordances in the labeling between clinicians. A study shows that the rate of agreement between clinicians in the same laboratory is about 95% [5]. In addition to the difficulties noted previously, we should not forget the time spent by this analysis (to label or score one night of 8h of sleep, per 30s epochs, a few hours of expert work is required on average).

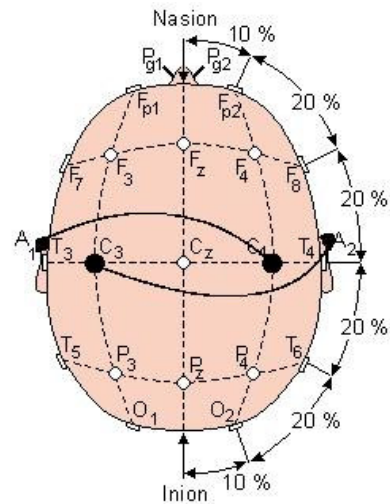

**Fig.1**: Electrodes placement in the standard 10-20 system.

## 2.2 Automatic Analysis

The development of automatic sleep analysis systems, intensified by the technological improvements in the micro-processing field, gave rise to numerical polygraphs. Currently, practically all the new numerical polygraphs are equipped with an automatic sleep analyzer more or less powerful [6][7][8]. The interest for these systems in the sleep study services is increasing for multiple reasons among which we can mention the most important:
- Considerable increase in the requests for recordings and the need for an automatic analysis system releasing the clinician from certain examination tasks;
- The growing number of the parameters intervening in a polygraphic recording and especially the need for quantifying and for classifying all these parameters;
- The interest to have a new vision of continuity and sleep architecture, not directly observable visually but only after signal processing like the activity with slow waves, the microstructure of the sleep, etc.
- The majority of the automatic analysis systems uses as reference the visual analysis criteria of Rechtschaffen and Kales (R&K) which remain the only consensus of sleep stages classification. Thus any measurement of performance of these systems is performed compared to the visual analysis.

An automatic analysis system can be described like an association of two parts: a data part and a treatments part. The data or the parameters are a representations of the physiological signals. This representation must be faithful with regard to the characteristics of the signal and must keep its fundamental properties while bringing a simplification, without great losses, in order to facilitate the following step: the treatment. This second part is composed by the treatment algorithms which aim to associate the data information to the sleep stage.

On the data part level, the major difficulty for these systems consists of the choice of modeling of the physiological signals. Indeed the choice of modeling influences considerably the performances of the system. Various techniques of analysis are used: amplitude analysis, period analysis, spectral analysis, etc [6][9][10]. Among these techniques the spectral analysis with Fast Fourier Transform (FFT) is mainly used. This choice can be explained by the fact that the visual analysis is based primarily on the detection of some waves with particular frequencies (Table 1).

On the treatment part level, research explored a broad range of traditional techniques or resulting from the artificial intelligence techniques. Among these techniques, we will be interested here in Artificial Neural Networks (ANN) which are the subject of our study and which will be presented in the following section. ANN are largely applied in neurophysiological fields like EEG analysis [11][12], somatosensory evoked potentials [13], analysis of vigilance [14], sleep analysis [15][16][17], etc.

As it was explained in the introduction part, the goal of this paper is to compare the performance of a human expert and an ANN in a simple labeling task, based on the same data and the same representation, in an experiment that we describe below.

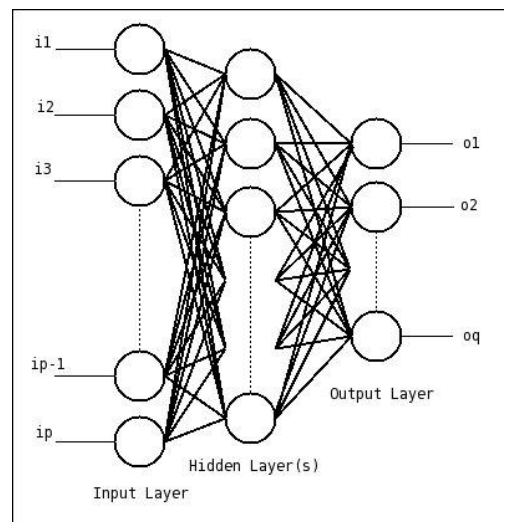

**Fig.2**: Multi Layer Perceptron (MLP) architecture.

## 3 Artificial Neural Networks

For our experiment, the Multi Layer Perceptron (MLP) [18] was selected as the most adequate tool. Among Artificial Neural Networks, MLPs are certainly the most frequently used, when a supervised training is required, i.e. when a matching must be learned between two kinds of data representing respectively the input and the output of the network. Between the layers of neurons designed to code these input and output signals, one or several intermediate hidden layers of arbitrary size can be added and allow to complexify the relation that can be learned between the input and the output layer. Generally connectivity is complete from one layer to the following layer (i.e. each neuron in one layer is connected to each neuron in the next layer), and a weight is assigned to each connection (Fig.2). Two phases of computation are then carried out for each step of processing, constituting the back-propagation learning algorithm [19]:
- The forward propagation of activity from the input to the output layer, giving the computed output that

can be compared with the desired output (which must be known for each example for supervised training). This comparison generally results in a measure of error.
- The backward propagation of error from the output layer to the input layer, giving the modification of each weight allowing for global error decrease. The weight modification rule is the core of the very classical back-propagation learning algorithm [19] and will not be described here.

These two phases of computation are performed for each example in a learning database during many epochs. Then, a validation database is used only in the forward way (computation of output activity and error) to estimate the MLP performance. Typically the learning error decreases as the number of learning iteration (epochs) increases and it is observed that the validation error reaches a minimum before growing up again as illustrated in Fig.3. In the statistical cross-validation framework, this process is repeated a given number of times on other learning and validation sets chosen randomly (ideally on all possible learning and validation sets obtained from the total set of example at disposal), and the performance of the network is computed as an average of all measured performances. Several internal parameters (e.g. learning rate) and architectural characteristics (e.g. number of hidden layers) of the network have to be specified. There exist several heuristics to explore these dimensions and the network which gives the best generalization, i.e. the smallest validation error, is selected as the final network.

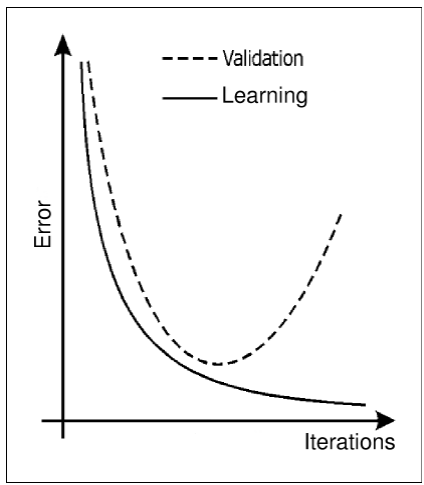

**Fig.3**: MLP learning curve.

## 4 Materials
### 4.1 Subject
This experiment used one sleep night recording of a 39 years old male adult. This subject was addressed to the service "Functional Explorations of the Nervous System" in the Sahloul Hospital, in Tunisia, for the suspicion of sleep disorder. The visual analysis as well as the report drawn up show that it is a healthy patient.

### 4.2 Data
The physiological signals collected on the subject are amplified, filtered, digitized (sampling frequency of 256 Hz) and finally recorded on hard disk in a format adapted to sleep polygraphy [20]. This polygraphy contains several signals among which the expert keeps for his analysis two EEG, one EOG and one EMG (Fig.4).

The two EEG derivations used are the centro-temporal C3-A2 left-right and the centro-temporal C4-A1 right-left (bold lines in Fig.1). During the visual analysis the expert bases his diagnosis primarily on the first derivation of EEG. To simplify our work we will adopt this derivation as single indicator for automatic classification. The Table 2 represents the details of the recording carried out as well as the structuration in sleep stages with the number of epochs, the duration, the percentage compared to the Total Time of Sleep (%TTS) and the percentage compared to the Total Time of Recording (%TTR). The TTS represents the cumulated duration of stages 1, 2, 3 and 4 in addition to the duration of the REM sleep. These values are obtained following the visual analysis per 30s epochs carried out by one expert. We give as an additional information the theoretical values of the percentages of the stages compared to the TTS (%TV-TTS) [3].

|          | Epochs | %TTS | %TV-TTS | %TTR |
|----------|--------|------|---------|------|
| Awake    | 67     | ---  | ---     | 6    |
| Stage 1  | 54     | 5    | < 10    | 5    |
| Stage 2  | 347    | 34   | ~ 50    | 31   |
| Stage 3  | 107    | 10   | ~ 10    | 10   |
| Stage 4  | 292    | 26   | ~ 10    | 26   |
| REM      | 233    | 21   | 20 to 25| 21   |
| Movement | 14     | ---  | ---     | 1    |
| TTS      | 1033   |      |         |      |
| TTR      | 1114   |      |         |      |

**Table 2** : Stage composition.

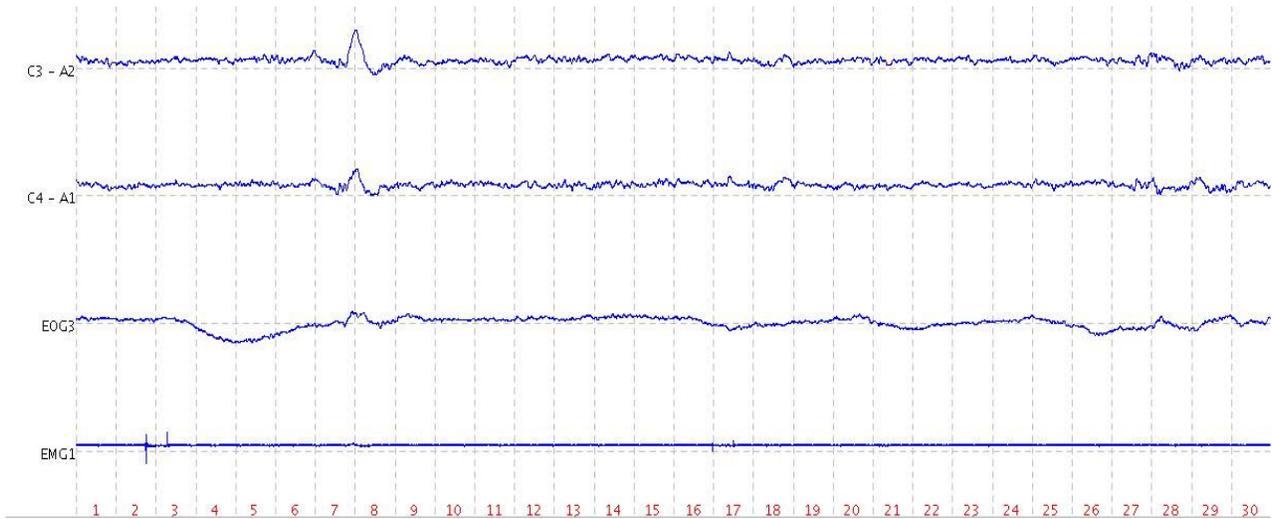

**Fig.4**: A sample screen showing physiological signals used for the visual analysis.

The visual analysis is carried out per 30s epochs, which yield 1114 samples for our recording. The distribution of these epochs in sleep stages is given in the Table 2. To model the selected EEG signal and to build our database we proceeded as follows (Fig.5):
- The signal is cut per 30s periods corresponding to the visual analysis epochs;
- For each portion of the signal we calculate the spectral power by the FFT. In the obtained spectrum we keep only the frequencies included in the interval ]0.5, 32Hz]. This interval corresponds to the field of variation of the physiological waves (lower than 32Hz) while eliminating the continuous component (close to 0Hz frequency);
- The spectrum is subdivided in five parts corresponding to the sleep bands (Delta = [0.5, 4Hz]; Theta = ]4, 8Hz]; Alpha = ]8, 12Hz]; Sigma = ]12, 16Hz]; and Beta = ]16, 32Hz]). For each band we calculate its relative spectral power (RSP) which is equal to the ratio of the band spectral power (BSP) by the total spectral power (TSP).

$$RSP_i = \frac{BSP_i}{TSP}, i \in \{Delta, Theta, Alpha, Sigma, Beta\} \quad (1)$$

All the functionalities and the operations needed to the physiological signals analysis and their modeling are carried out by a software which we developed for this effect[1]. Thus each 30s epoch will be represented in our database by the five values of the RSP to which we associate a label representing the sleep stage. In this database we did not integrate the epochs scored as movement considering their low number and their poor importance in our study. Finally our corpus will be composed with 1100 samples distributed in the 5 sleep stages and the awake.

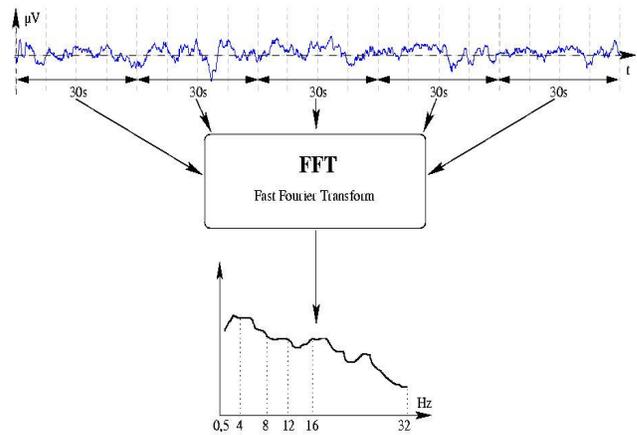

**Fig.5**: EEG spectral modeling.

## 5 Results and discussion
### 5.1 Results
The experiment was carried out using a Multi-Layer Perceptron (MLP). The number of input neurons of the ANN used is fixed to 5 corresponding to the 5 values of the RSP calculated on the EEG derivation (Equation 1). The number of output neurons is fixed to 6, a neuron for each sleep stage in addition to the awake.

---

[1] jEDF home page: http://www.loria.fr/~kerkeni/jEDF.html

The number of the hidden neurons is obtained after a study of several configurations, during the training process. The best success rate obtained among the various configurations tested is 76%. This rate is obtained with an ANN composed by 5 input neurons, 6 neurons in the hidden layer and 6 output neurons. Simulations are carried out by cross validation on ten random selected sets. The confusion matrix (Table 3), result of the optimal network classification, shows that:
1. Awake, stage 2, stage 4 and the REM sleep are well classified;
2. Stage 1 is not recognized and it is mainly confused with the REM sleep then with stage 2;
3. Stage 3 is slightly recognized and it is confused with stage 4.

| as → | Awake | S1 | S2 | S3 | S4 | REM | Success |
|---|---|---|---|---|---|---|---|
| **Awake** | **59** | 0 | 0 | 0 | 5 | 3 | **88%** |
| **S1** | 11 | **0** | 17 | 0 | 2 | 24 | 0% |
| **S2** | 3 | 1 | **291** | 0 | 21 | 31 | **84%** |
| **S3** | 0 | 0 | 39 | **3** | 52 | 13 | 3% |
| **S4** | 1 | 0 | 9 | 2 | **278** | 2 | **95%** |
| **REM** | 7 | 0 | 16 | 0 | 6 | **204** | **88%** |

Table 3: Confusion matrix.

These results are also reproduced, with some differences in the numerical values, with the second EEG derivation and the same with the combinations of the two EEG derivations (10 input parameters, 5 parameters for each derivation). The combination of the two derivations does not make significant improvement to the level of the total success which passes from 76% to 77%.

## 5.2 Discussion

The high recognition rate of stage 2, stage 4 and REM is explained by the fact that they are well represented in the training database, and this improves their recognition by the ANN in spite of the resemblance, on the spectral level, between stage 2 and REM. In spite of the poor representation of the awake stage, it is well recognized because it presents a different spectral composition, as compared to the other stages: the Alpha waves are present only in the awake stage (as can be seen in Table 1).

Stage 1 is not recognized because it is too slightly represented in the database and it has the same spectral composition as stage 2 and REM which are strongly represented, and this can certainly explain the observed confusion.

Stage 3 is confused with stage 4 because they have the same spectral composition, i.e. predominance of the Delta waves. Even the rule used for visual classification is based on a subjective measure for discrimination between these two stages. Indeed, it is sufficient that the Delta waves occupy more than 50% of the epoch duration to score it in stage 4 instead of stage 3.

As an indication, we give below the spectral composition in sleep bands of stage 3 and stage 4 examples present in our database (Fig.6). This can contribute to explain why these stages can be confused if they are labeled only on the basis of spectral composition. The same illustration could be displayed for other confused stages listed above and would also contribute to state that a pure spectral analysis is not sufficient to fully differentiate between the stages of sleep.

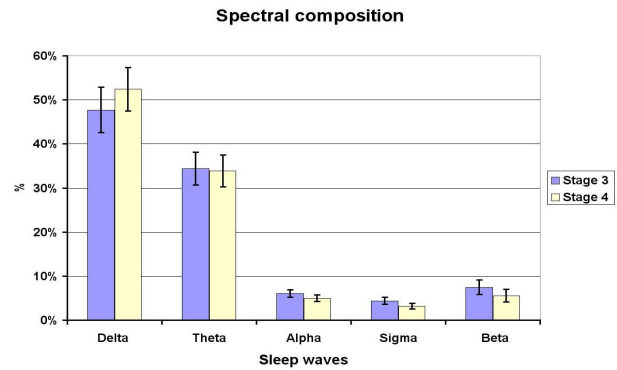

Fig.6: Spectral composition of stage 3 and 4 present in the data base.

## 6 Conclusion

The objective of this work was to advance our experience in producing a decision-making tool for sleep analysis based on Artificial Neural Networks. In this paper we presented one of our steps to better understand how we can compare human expertise and ANN decision. The presented configuration gives a 76% rate of agreement for the identification of the 6 stages of sleep from restricted EEG input data. Equivalent works in the literature, using ANN for a classification in 6 classes, give results which vary between 61 and 80% [11].

Even if in [1] we have explained that disagreement does not always mean that the ANN is wrong, we are conscious, in such a simple classification stage (sleep without noise nor movement), that these results mainly show the limit of the FFT modeling adopted in our study and this results in the confusion between stages equivalent at the spectral level. Another choice of parameters and/or the addition of other parameters resulting from other modeling techniques like the detection of the graphical-elements and the integration of the other physiological signals might be able to improve the obtained results. It is what we are studying in ongoing work.

We are also trying to more precisely understand human expertise and particularly its structure. Indeed, it appears that all rules are not equivalent, from a priority point of view but also for the time span that they imply, whereas ANN are known to difficultly integrate structured data and knowledge. Embedding such typically human characteristics in an automatic neuronal processing would be an extremely interesting property to give to our physiological signal analysis systems.

*Acknowledgment:*
The authors wish to thank Mrs. Jaidane Nadia, service of "Functional Explorations of the Nervous System" in the Sahloul Hospital, in Tunisia, for her precious contribution and assistance in the selection and the data-gathering used for this paper.
This work was performed in the framework of a STIC project (funded by INRIA and Tunisian Universities) and with the help of the Lorraine Region.